\documentclass{article}
\usepackage{spconf,amsmath,graphicx, todonotes, float, multirow}

\usepackage{enumitem}
\usepackage{algpseudocode, algorithm}
\setlist{nosep, leftmargin=14pt}

\usepackage{mwe} 

\title{Detecting and refurbishing ground truth errors during training of deep learning-based echocardiography segmentation models}
\name{Iman Islam, Bram Ruijsink, Andrew J. Reader, Andrew P. King}
\address{School of Biomedical Engineering and Imaging Sciences, King's College London, London, UK}
\begin{document}
%
\maketitle
\begin{abstract}
Deep learning-based medical image segmentation typically relies on ground truth (GT) labels obtained through manual annotation, but these can be prone to random errors or systematic biases. This study examines the robustness of deep learning models to such errors in echocardiography (echo) segmentation and evaluates a novel strategy for detecting and refurbishing erroneous labels during model training. Using the CAMUS dataset, we simulate three error types, then compare a loss-based GT label error detection method with one based on Variance of Gradients (VOG). We also propose a pseudo-labelling approach to refurbish suspected erroneous GT labels. We assess the performance of our proposed approach under varying error levels. Results show that VOG proved highly effective in flagging erroneous GT labels during training. However, a standard U-Net maintained strong performance under random label errors and moderate levels of systematic errors (up to 50\%). The detection and refurbishment approach improved performance, particularly under high-error conditions.
\end{abstract}
\section{Introduction}
\label{sec:intro}
Deep learning (DL) models have been widely proposed for automating the segmentation of cardiac structures from echocardiography (echo) images \cite{leclerc_deep_2019,tromp_automated_2022}. Subsequently, these segmentations are often used to calculate functional biomarkers such as ejection fraction and therefore their reliability is of crucial importance in patient management \cite{puyol-anton_ai-enabled_2022, ghorbani_deep_2020}.

The training of DL segmentation models requires access to (preferably) large datasets with associated annotations, or ground truth (GT) labels of the structures of interest. These GT labels are often treated as a `gold standard'; however, they are typically created by humans who are prone to making mistakes, particularly in complex and laborious tasks such as medical image segmentation. The assumption is always that GT labels are correct, but in reality minor differences due to inter-annotator variability \cite{leclerc_deep_2019} and more severe annotation errors are common in medical imaging \cite{mariscal2023artificial}. The impact that such errors can have on model performance when they are included in the training data has been mostly overlooked in DL-based segmentation of medical images. 

\subsection{Aims and Contributions}
First, this study aims to investigate the robustness of a state-of-the-art (SOTA) DL segmentation model to errors in the GT labels used in training. A range of synthetic but realistic errors are introduced into the training data with varying frequencies and their impact on model performance is measured.

Second, we propose an automated technique to detect potentially problematic GT labels and a method to `refurbish' them during training. The necessity and utility of this approach is investigated.

\subsection{Related Work}

Most work to date on identifying and dealing with GT label errors has been in classification tasks, and there has been limited work in segmentation tasks. Segmentation introduces unique challenges that are not present in classification tasks, such as spatial continuity and boundary sensitivity.

In general, research into GT label errors mainly considers three tasks: (i) identifying which GT labels are noisy, (ii) determining the optimal stage in the training process to intervene and (iii) selecting an appropriate method to `clean' or `refurbish' them.

The majority of methods for detecting GT label errors when training classification models assume that samples with persistently high training losses are likely to be mislabelled. Therefore, in order to select the erroneous labels, these works have considered the loss distribution of samples \cite{song2019selfie, li2020dividemix}. Some papers have also proposed considering the model’s prediction confidence, and judged low confidence to indicate errors in the GT label \cite{huang2020self}. Additionally, a recent paper suggested considering the variance of gradients (VOG) for individual samples, assuming that large changes in gradients are indicative of errors in the GT label \cite{khanal2024active}.

In order to reduce the impact of erroneous labels, several label refurbishment techniques have been proposed. One common approach is pseudo-labelling, where potentially incorrect GT labels are replaced with the current model’s prediction during training \cite{song2019selfie, huang2020self, li2020dividemix}. Another approach is reweighting losses, in which samples suspected to be mislabelled are assigned lower weights during optimisation \cite{chen2021beyond}. In addition, sample filtering methods discard samples during training so that the model learns only from clean samples \cite{han2018co}.
These label refurbishment techniques have been implemented from the start of training \cite{huang2020self, chen2021beyond, li2020dividemix} or after a warm-up period \cite{song2019selfie}.

Fewer studies have explicitly addressed GT label errors in segmentation tasks.
One approach has involved using model uncertainty to locate label regions that are likely to be incorrect to inform pixel-level relabelling 
\cite{redekop2021uncertainty}. Alternatively, an iterative refinement pipeline has been proposed in which the initial segmentation network generated improved annotations that were used to supervise a second stage network to reduce the reliance on imperfect original labels over time \cite{xue2020cascaded}. Another approach included applying adaptive early-learning to emphasize clean samples during training \cite{liu2022adaptive}. Finally, self-relabelling techniques have been proposed to refine segmentation labels dynamically based on historical predictions \cite{li2022self}.

Currently, significant gaps remain in the literature concerning the impact of GT label errors on SOTA segmentation models. First, prior work has predominantly been restricted to binary segmentation scenarios, ignoring the common task of multi-class segmentation. Second, although there are some methods to mitigate the effects of GT label errors, a comprehensive evaluation of the robustness of models to different types of errors is lacking. This lack of evaluation limits our understanding of how well these models handle real-world annotation errors. Finally, the impact of incorrect GT segmentations in echo images remains unexplored. Given the inherent challenges of echo image segmentation, such as variability in anatomical structures and poor image quality, addressing these limitations is essential to improve model reliability and trust in real-world clinical settings.

\section{Materials and Methods}
\label{sec:methods}
\subsection{Dataset}

We used the CAMUS dataset \cite{leclerc_deep_2019}, which contains 2D echo images with manual labellings of the left ventricle (LV), left ventricular myocardium (LVM) and left atrium (LA). Specifically, we selected the apical four-chamber (A4C) view images at end-diastole (ED), resulting in a total of 500 annotated samples. 

\subsection{GT Label Error Generation}
\label{sec:errors}
\begin{figure}[h]
    \centering
    \includegraphics[width=3in]{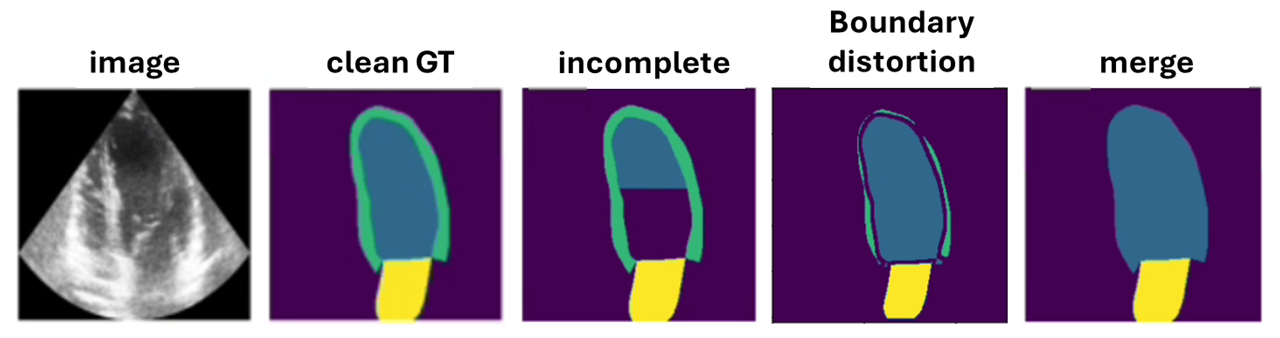}
    \caption{A sample image and GT label with examples of the three types of synthetic errors: incomplete label, boundary distortion and merged label. Blue=LV, Green=LVM, Yellow=LA}
    \label{fig:egGTerrors}
\end{figure}

To evaluate the impact of GT label errors, we introduced synthetic but realistic errors into a random subset of the training and validation data. The specific error types used in this study are often found in real-life echo datasets and are summarised below (examples shown in Fig. \ref{fig:egGTerrors}):

\begin{itemize} \item[(i)] \emph{Incomplete label error}: Part of a structure is left unlabelled (i.e., as background), simulating interruptions or incomplete annotations.
\item[(ii)] \emph{Boundary distortion error}: Labels are dilated or eroded, mimicking confusion at boundaries and different annotation styles.
\item[(iii)] \emph{Merged labels error}: One structure is incorrectly labelled as another already labelled structure, simulating unchecked label application.
\end{itemize}

\noindent We examined two categories of error: \textbf{random} errors - arbitrary mistakes simulating inattentive annotation where there is an equal chance of every structure being affected, and \textbf{systematic} errors - consistent mislabelling of specific structure(s). Systematic errors were applied as follows: (i) for incomplete labels, the LV was consistently partially annotated; (ii) for boundary distortion, the labels were consistently eroded; and (iii) for merged labels, the LVM was relabelled and merged with the LV class.

\subsection{Baseline Segmentation Model}

A U-Net \cite{ronneberger_u-net_2015} was used as the baseline model to segment the LV, LVM, and LA. Models were trained with cross-entropy loss using the Adam optimizer (learning rate 0.001, batch size 4) for 100 epochs. Hyperparameters were selected via a grid search based on validation performance. Data were split 80\%/10\%/10\% into training, validation, and test sets. The model from the epoch with the best validation foreground Dice score was used for test evaluation.

\subsection{GT Label Error Detection and Refurbishment Pipeline}

A summary of the GT label error detection and refurbishment pipeline is provided in Algorithm \ref{alg:label_ref}. Details of the detection and refurbishment steps are provided below.

\subsubsection{Label Error Detection}
To detect GT label errors, we use the VOG method which measures gradient variability of each training sample's activations over several epochs. Samples with high gradient variability are likely mislabeled or noisy. We adopt the VOG approach by Khanal et al.
\cite{khanal2024active}, as defined in Equation \ref{eq:vog}. 

\begin{equation}
\text{VOG}_{ij} = \frac{1}{D} \sum_{d=1}^{D} \sqrt{ \frac{1}{t} \sum_{e=j-t}^{j} \left(S_{ie} - \mu_i \right)^2 }
\label{eq:vog}
\end{equation}

\noindent where $D$ is the dimension of the gradient vector, \(t=5\) is the number of previous epochs used to compute the variance, $S_{ij}$ is the gradient vector of the activation for a sample $x_i$ at epoch $j$, and \(\mu_i = \frac{1}{t} \sum_{e=j-t}^{j} S_{ie}\) is the mean gradient vector over the last \(t\) epochs.

To identify potential GT label errors, we apply a threshold based on the upper bound of the interquartile range (IQR): samples with VOG scores greater than $Q_3 + 1.5 \times \text{IQR}$ are flagged as outliers ($Q_3$ is the third quartile value). This robust criterion is commonly used in outlier detection and avoids reliance on arbitrary cutoffs.

\subsubsection{Refurbishment}

Adapted from SELFIE \cite{song2019selfie}, we perform label refurbishment every 5 epochs following a 10-epoch warm-up. 
Each refurbishment replaces the training label with a pseudo-label obtained by averaging the model's predictions over the preceding 5 epochs (Equation \ref{eq:selfie}).


\begin{equation}
\hat{y}_i = \frac{1}{5} \sum_{e=E-4}^{E} p_i^{(e)}
\label{eq:selfie}
\end{equation}

\noindent
where \( \hat{y}_i \) is the refined pseudo-label for sample \( i \), \( p_i^{(e)} \) is the model's prediction for sample \( i \) at epoch \( e \), and \( E \) is the current epoch. 

\begin{algorithm}[h]
\caption{Iterative VOG based Label Refurbishment}
\label{alg:label_ref}
\begin{algorithmic}[1]
\Require $S$: training set data,  $e$: $\text{epochs}$, $T$: warm-up period, $t$: iteration interval
\Ensure Model parameters, $R$: refurbished samples
\For{each epoch $e$ in \text{epochs}}
    \For{each batch $B$ in $S$}
        \State Train model on $B$
        \State Record gradients for every sample
    \EndFor
    \If{$e > T$ and $e \bmod t = 0$}
        \State Calculate VOG for every sample
        \For{samples with $VOG > Q3 + 1.5 \times IQR$}
            \State Apply label refurbishment method
            \State Update GT segmentations in $D$
        \EndFor
    \EndIf
\EndFor
\end{algorithmic}
\end{algorithm}

\section{Experiments and Results}
\label{sec:experiments}
The purpose of these experiments is to evaluate (i) the impact of GT label errors on segmentation performance and (ii) the effectiveness of the GT label error detection and refurbishment method.

\subsection{Experiment 1: GT Label Error Detection}

To evaluate the effectiveness of our method for identifying GT label errors, we conducted an experiment in which we apply random label errors to 25\% of the training and validation sets. This was done for the three label errors detailed in Section \ref{sec:errors}. We compared our VOG based strategy with the more commonly used loss-based approach (i.e. based upon the training loss rather than VOG, with the same outlier detection method). We tested two different proportions of GT label errors. Table \ref{tab:comp} summarises the results. In most cases, the VOG method consistently outperformed the loss-based approach, achieving higher accuracy, sensitivity and specificity in identifying GT label errors. Therefore, we use the VOG based strategy in our subsequent experiments.

\begin{table}[h]
\centering
\caption{Experiment 1 - Comparison of methods for identifying erroneous GT labels.}
\label{tab:comp}
\resizebox{\columnwidth}{!}{%
\begin{tabular}{|c|cccccc|}
\hline
 &
  \multicolumn{2}{c}{\textbf{\begin{tabular}[c]{@{}c@{}}Incomplete\\ labels\end{tabular}}} &
  \multicolumn{2}{c}{\textbf{\begin{tabular}[c]{@{}c@{}}Boundary\\ distortion\end{tabular}}} &
  \multicolumn{2}{c|}{\textbf{\begin{tabular}[c]{@{}c@{}}Merged\\ labels\end{tabular}}} \\ \cline{2-7} 
                     & \textbf{VOG} & \multicolumn{1}{c|}{\textbf{Loss}} & \textbf{VOG} & \multicolumn{1}{c|}{\textbf{Loss}} & \textbf{VOG} & \textbf{Loss} \\ \hline
\textbf{Accuracy}    & 0.90         & \multicolumn{1}{c|}{0.81}          & 0.99         & \multicolumn{1}{c|}{0.95}          & 0.99         & 0.90          \\
\textbf{Sensitivity} & 0.86         & \multicolumn{1}{c|}{0.32}          & 0.97         & \multicolumn{1}{c|}{0.87}          & 1            & 1             \\
\textbf{Specificity} & 0.91         & \multicolumn{1}{c|}{0.97}          & 0.99         & \multicolumn{1}{c|}{0.97}          & 0.99         & 0.87          \\ \hline
\end{tabular}%
}
\end{table}

\subsection{Experiment 2: Label Refurbishment}

To assess the refurbishment strategy’s effectiveness in correcting GT label errors, we compared Dice scores of refurbished labels before and after refurbishment, using the original GT labels as reference. This was tested under random and systematic label errors. As shown in Table \ref{tab:refurbish}, refurbishment consistently improved Dice scores, especially when random errors were below 50\% and systematic errors below 25\%.


\begin{table}[h]
\centering
\caption{Experiment 2 - Evaluation of refurbishment method. Dice scores for the erroneous dataset before and after refurbishment.}
\label{tab:refurbish}
\resizebox{\columnwidth}{!}{%
\begin{tabular}{|cccccc|}
\hline
 &
   &
   &
  \textbf{\begin{tabular}[c]{@{}c@{}}Incomplete\\ labels\end{tabular}} &
  \textbf{\begin{tabular}[c]{@{}c@{}}Boundary\\ distortion\end{tabular}} &
  \textbf{\begin{tabular}[c]{@{}c@{}}Merge\\ labels\end{tabular}} \\ \hline
\multicolumn{1}{|c|}{\multirow{4}{*}{\rotatebox[origin=c]{90}{\textbf{Random}}}}     & \textbf{}       & \textbf{before} & $0.85 \pm 0.17$ & $0.68 \pm 0.06$ & $0.53 \pm 0.45$ \\ \cline{2-6} 
\multicolumn{1}{|c|}{}                                                               & \textbf{$12.5\%$} & \textbf{after}  & $0.95 \pm 0.03$ & $0.84 \pm 0.06$ & $0.93 \pm 0.04$ \\
\multicolumn{1}{|c|}{}                                                              & \textbf{$25\%$}   & \textbf{after}  & $0.96 \pm 0.04$ & $0.75 \pm 0.13$ & $0.93 \pm 0.04$ \\
\multicolumn{1}{|c|}{}                                                               & \textbf{$50\%$}   & \textbf{after}  & $0.90 \pm 0.14$ & $0.70 \pm 0.10$ & $0.69 \pm 0.37$ \\ \hline
\multicolumn{1}{|c|}{\multirow{4}{*}{\rotatebox[origin=c]{90}{\textbf{Systematic}}}} & \textbf{}       & \textbf{before} & $0.68 \pm 0.03$ & $0.52 \pm 0.11$ & $0.74 \pm 0.04$ \\ \cline{2-6} 
\multicolumn{1}{|c|}{}                                                               & \textbf{$12.5\%$} & \textbf{after}  & $0.94 \pm 0.04$ & $0.87 \pm 0.04$ & $0.94 \pm 0.03$ \\
\multicolumn{1}{|c|}{}                                                               & \textbf{$25\%$}   & \textbf{after}  & $0.75 \pm 0.10$ & $0.72 \pm 0.15$ & $0.79 \pm 0.09$ \\
\multicolumn{1}{|c|}{}                                                               & \textbf{$50\%$}   & \textbf{after}  & $0.69 \pm 0.06$ & $0.54 \pm 0.13$ & $0.77 \pm 0.08$ \\ \hline
\end{tabular}%
}
\end{table}

\subsection{Experiment 3: Complete Pipeline}

\begin{figure}[h]
    \centering
    \includegraphics[width=3.5in]{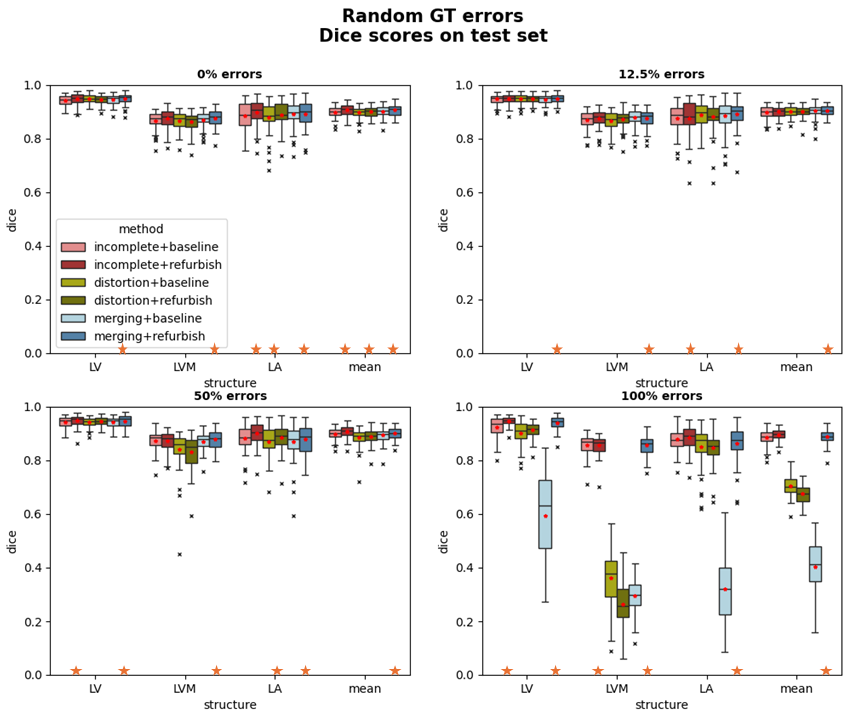}
    \caption{Experiment 3 - Random GT label errors. Test set results when training with random GT label errors. Box plots display the Dice coefficients for each segmented structure and the overall mean for each model. The orange stars represent cases where the refurbished model performs better than the baseline model with a statistically significant difference by a Wilcoxon signed rank test (p-value\textless{}0.05).}
    \label{fig:random}
\end{figure}

To evaluate model robustness to random GT label errors, we trained both a baseline U-Net and a model incorporating our GT label error detection and refurbishment strategy. Synthetic random segmentation errors were introduced into the training and validation sets at varying proportions, i.e., for each of the three error types (incomplete label, boundary distortion and merged labels), random class(es) were chosen for each affected sample. Results are shown in Fig. \ref{fig:random}.

As the proportion of incorrect labels increased, segmentation performance gradually declined. Surprisingly, despite the introduction of GT label errors, the baseline U-Net maintained relatively strong performance. The refurbishment strategy resulted in some improvements over the baseline, especially at higher proportions of errors.

\begin{figure}[h]
    \centering
    \includegraphics[width=3.5in]{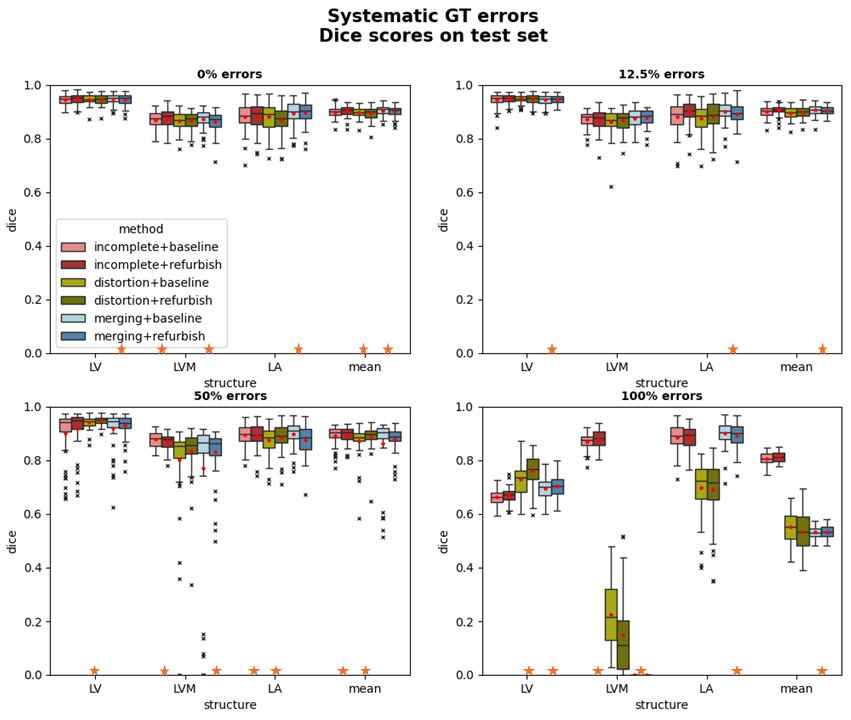}
    \caption{Experiment 3 - Systematic GT errors. Test set results when training with systematic GT label errors. Box plots display the Dice coefficients for each segmented structure and the overall mean for each model. The orange stars represent cases where the refurbished model performs better than the baseline model with a statistically significant difference by a Wilcoxon signed rank test (p-value\textless{}0.05).}
    \label{fig:systematic}
\end{figure}

We repeated the previous experiment to assess model performance under systematic GT label errors. Results are shown in Fig. \ref{fig:systematic}.

Systematic errors led to a more pronounced degradation in segmentation quality compared to random errors. In the boundary distortion and merged labels scenarios, segmentation performance deteriorated significantly, with a higher incidence of outlier cases. With incomplete labels, the baseline model continued to perform reasonably well, again showing good robustness. Across all systematic error types, the refurbishment strategy offered small gains over the baseline at low levels of GT label errors.

\section{Discussion and Conclusion}
\label{sec:discussion}
Across all experiments, we observed that the baseline segmentation model was surprisingly robust to a wide range of GT label errors. Random label errors had surprisingly little effect on final performance.
For systematic errors, refurbishment provided some improvement, especially at moderate levels of corruption. This suggests that when the GT label errors follow structured patterns, error detection and refurbishment strategies may have a role to play. 



In practice, annotation errors typically affect less than 10\% of data, making our high-error experiments a worst-case scenario. To our knowledge, the observed robustness of a naively trained U-Net under such conditions has not been previously reported and we believe that these findings offer a valuable contribution to the field.

We also found that VOG reliably identified corrupted labels across error types and proportions, making it a practical, lightweight method for flagging incorrect GT labels without retraining or architectural changes.

Taken together, these results highlight a promising degree of resilience in segmentation models and point toward selective, rather than routine, use of label refurbishment and correction strategies in clinical practice.





\section{Compliance with ethical standards}
This research study was conducted retrospectively using human subject data made available in open access \cite{leclerc_deep_2019}. Ethical approval was not required as confirmed by the license attached with the open access data.

\section{Acknowledgments}
\label{sec:acknowledgments}
We would like to acknowledge funding from the EPSRC Centre for Doctoral Training in Medical Imaging (EP/L015226/1).

\bibliographystyle{IEEEbib}
\bibliography{references.bib}

\end{document}